\documentclass[letterpaper, 10 pt, conference]{ieeeconf}

\usepackage[english]{babel}
\usepackage{inputenc}
\usepackage{amsmath}
\usepackage{graphicx}
\usepackage{caption}
\usepackage{subcaption}
\usepackage{booktabs}
\usepackage{mathtools}
\usepackage{xcolor}
\usepackage{amsfonts}
\usepackage{float}
\usepackage{url}
\usepackage[nolist,nohyperlinks]{acronym}
\IEEEoverridecommandlockouts
\newcommand{\reffig}[1]{Fig.~\ref{#1}}
\renewcommand{\refeq}[1]{(\ref{#1})}
\newcommand{\reftab}[1]{Table~\ref{#1}}
\newcommand{\refsec}[1]{Section~\ref{#1}}

\DeclareMathOperator*{\argmin}{\arg\!\min}

\renewcommand{\vec}[1]{\mathbf{#1}}
\newcommand{\mat}[1]{\mathbf{#1}}

\newcommand{\ra}[1]{\renewcommand{\arraystretch}{#1}}

\usepackage{pgfplots}
\pgfplotsset{compat=newest}
\usetikzlibrary{plotmarks}
\usepgfplotslibrary{patchplots}
\usepackage{grffile}
\usepackage{amsmath}
\newlength\figureheight
\newlength\figurewidth

\usepackage{microtype}

\title{C-blox: A Scalable and Consistent TSDF-based Dense Mapping Approach}
\author{Alexander Millane, Zachary Taylor, Helen Oleynikova, Juan Nieto, Roland Siegwart, C\'{e}sar Cadena\\
 Autonomous Systems Lab, ETH  Z{\"u}rich%
\thanks{This research was funded by the National Center of Competence in Research (NCCR) Robotics through the Swiss National Science Foundation.}}

\begin{document}
\maketitle

\begin{acronym}
\acro{TSDF}{Truncated Signed Distance Field}
\acro{AR}{Augmented Reality}
\acro{BA}{Bundle Adjustment}
\acro{AMD}{Approximate Minimum Degree}
\acro{SGBM}{Semi-Global Block Matching}
\acro{MAV}{Micro Aerial Vehicle}
\acro{RMSE}{Root-Mean-Square Error}
\acro{VI}{Visual-Inertial}
\acro{SLAM}{Simultaneous Localization And Mapping}
\end{acronym}

\begin{abstract}
In many applications, maintaining a consistent dense map of the environment is key to enabling robotic platforms to perform higher level decision making. Several works have addressed the challenge of creating precise dense 3D maps from visual sensors providing depth information. However, during operation over longer missions, reconstructions can easily become inconsistent due to accumulated camera tracking error and delayed loop closure. Without explicitly addressing the problem of map consistency, recovery from such distortions tends to be difficult. We present a novel system for dense 3D mapping which addresses the challenge of building consistent maps while dealing with scalability. Central to our approach is the representation of the environment as a collection of overlapping \ac{TSDF} subvolumes. These subvolumes are localized through feature-based camera tracking and bundle adjustment. Our main contribution is a pipeline for identifying stable regions in the map, and to fuse the contributing subvolumes. This approach allows us to reduce map growth while still maintaining consistency. We demonstrate the proposed system on a publicly available dataset and simulation engine, and demonstrate the efficacy of the proposed approach for building consistent and scalable maps. Finally we demonstrate our approach running in real-time on-board a lightweight \ac{MAV}.
\end{abstract}

\section{Introduction}
\label{sec:introduction}

Vision-based perception systems are increasingly being deployed for use on robotic platforms that operate in unstructured environments or without access to reliable GPS coverage \cite{burri2015real}. In addition to offering a sensing solution that does not depend on any external infrastructure, the benefits of such systems include their low weight, low cost and the richness of the data they provide.

Key competencies towards achieving high level tasks for robotic systems utilizing vision, are building an internal representation of the environment and localizing within it, known as \ac{SLAM}. The \ac{SLAM} problem, has been a focus of robotics research for the last three decades. Most successful \ac{SLAM} systems utilizing visual data simplify the problem by converting incoming images to a set of visual features, before estimating the camera motion and the map as a function of only these feature observations \cite{mur2016orb}. A summary of past and present \ac{SLAM} systems can be found in \cite{cadena2016past}.


\begin{figure}[tb]
    \centering
    \begin{subfigure}[b]{0.20\textwidth}
    \includegraphics[trim={0cm 0cm 0cm 0cm},clip,width=\columnwidth]{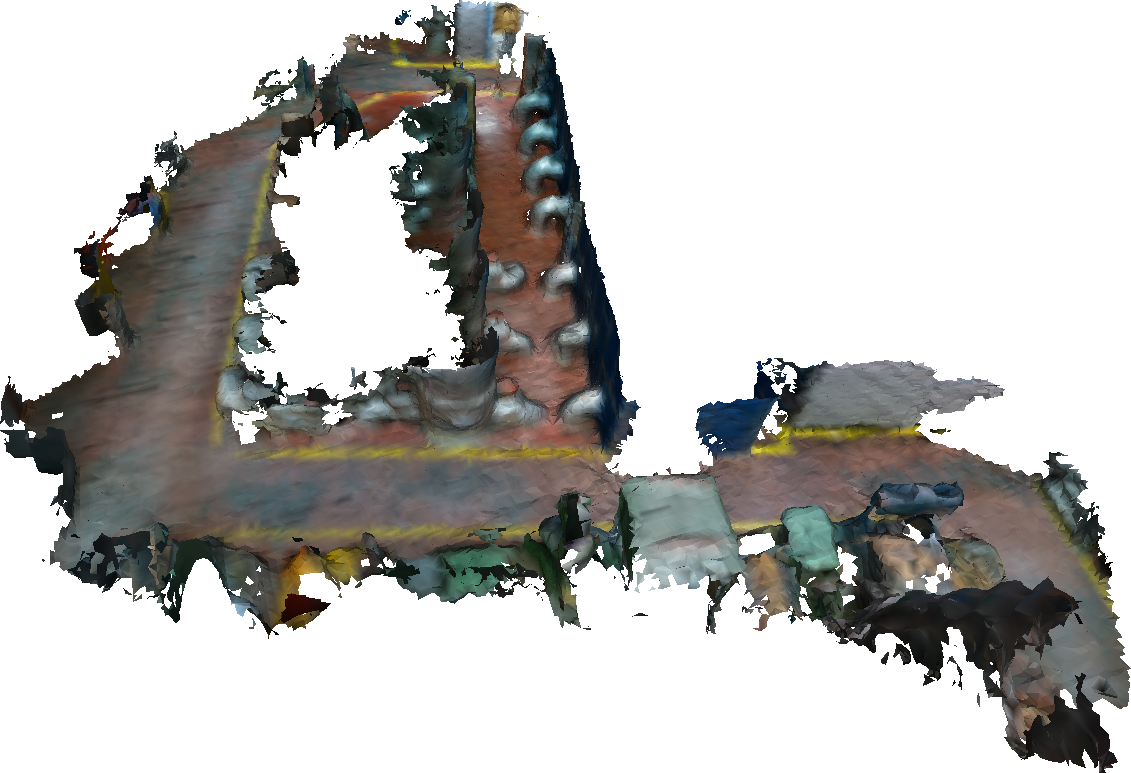}
        \caption{}
        \label{fig:cover_loop}
    \end{subfigure}
    \begin{subfigure}[b]{0.20\textwidth}
    \includegraphics[trim={0cm 0cm 0cm 0cm},clip,width=\columnwidth]{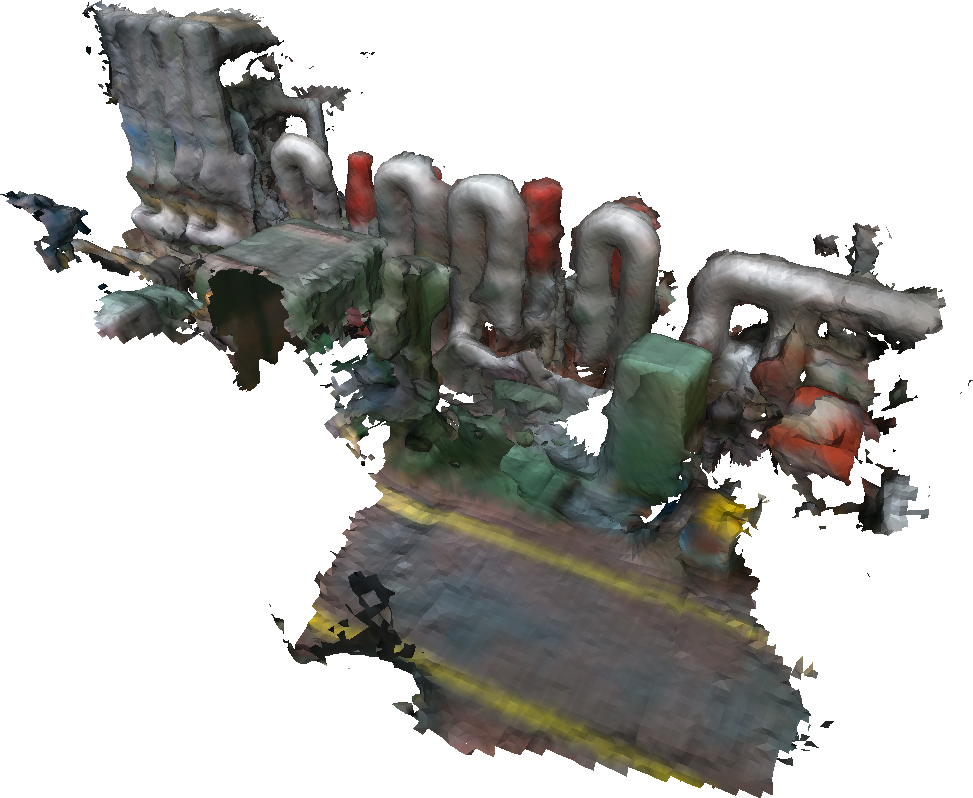}
        \caption{}
        \label{fig:cover_pipes}
    \end{subfigure}
    \begin{subfigure}[b]{0.20\textwidth}
    \includegraphics[trim={0cm 0cm 0cm 0cm},clip,width=\columnwidth]{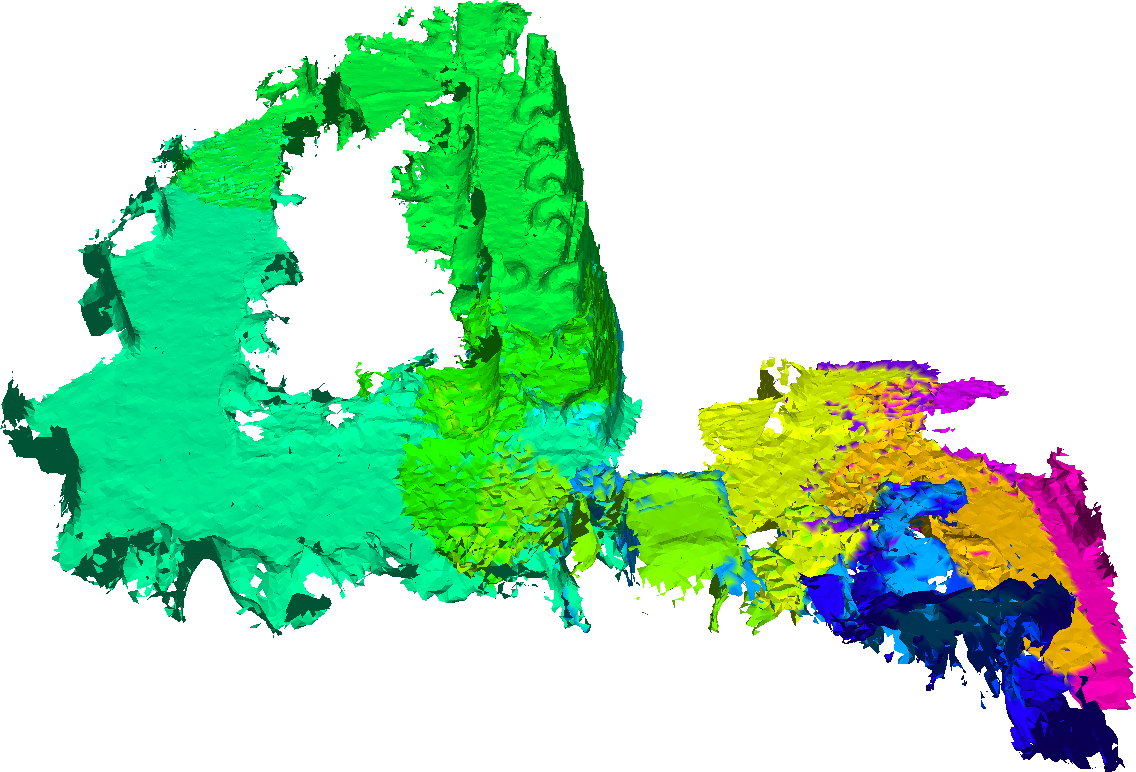}
        \caption{}
        \label{fig:cover_separated}
    \end{subfigure}
    \begin{subfigure}[b]{0.20\textwidth}
    \includegraphics[trim={0cm 0cm 0cm 0cm},clip,width=\columnwidth]{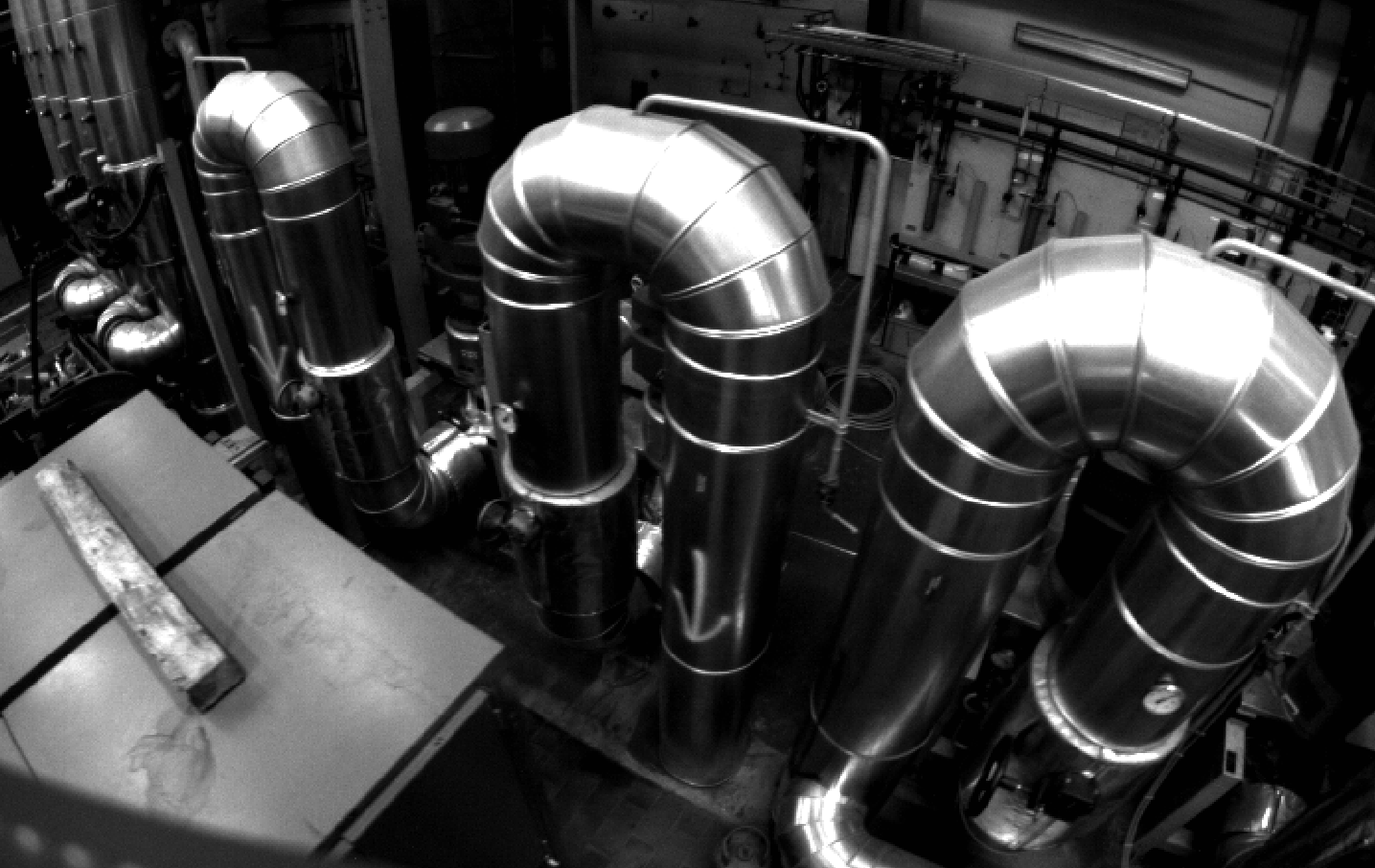}
        \caption{}
        \label{fig:cover_cam}
    \end{subfigure}
    \caption{Reconstructions resulting from 2 MAV flights in an indoor industrial area. The agent trajectories include several wide-baseline loop closures and frequent re-observation of similar parts of the scene. Figures (\subref{fig:cover_loop}) and (\subref{fig:cover_pipes}) show the final reconstruction from the two flights, (\subref{fig:cover_separated}) shows the breakdown of the flight 1 map into its composite submaps, and (\subref{fig:cover_cam}) shows a view from the onboard camera onto the pipe-like structures visible in the reconstruction (\subref{fig:cover_pipes}).}
    \label{fig:cover}
\end{figure}

While feature-based systems have proven themselves effective for localization, a map comprised of a sparse collection of 3D points is insufficient for tasks such as planning or environmental interaction. Recently, the commodification of depth cameras has seen impressive results produced in the field of dense 3D reconstruction from visual sensors \cite{newcombe2011kinectfusion}, \cite{whelan2012kintinuous}. The techniques used by these works are now making their way into robotics applications \cite{lin2017autonomous}, \cite{wagner2014graph}. However, the shortcomings of \ac{SLAM} systems mean that they suffer from imperfect odometry, resulting in accumulating pose drift, and delayed loop closures. Using a \ac{SLAM} system for camera tracking in an online dense reconstruction pipeline can therefore produce inconsistent reconstructions if these effects are not handled explicitly.

This paper introduces \emph{c-blox}\footnote{\url{https://github.com/ethz-asl/c-blox}}, a novel dense mapping system which addresses the challenge of maintaining map consistency and scalability. Central to our system is the representation of the observed scene as a collection of overlapping \ac{TSDF} subvolumes\footnote{We use the term subvolume to refer to 3D dense submaps, also termed \textit{patches} in \cite{howard2004multi} and \cite{henry2013patch}.}. In our approach we create new subvolumes early and often to limit the amount of intra-volume distortion. However, the approach of continuously creating subvolumes has the effect of building a map that grows over time and without bound. We therefore limit the resulting map growth by performing subvolume fusion, where doing so has a high probability of producing consistent results. We first follow a landmark covisibility \cite{mei2010closing} based approach to identify subvolumes containing potentially redundant views of the environment. In a second stage, we determine if these candidate subvolumes are well localized with respect to one another by extracting a measure of the relative localization certainty of the candidates. This certainty measure is extracted from the co-constructed sparse map to which the subvolumes are attached. In summary the contributions of our system are,
\begin{itemize}
\item a systematic approach for maintaining consistency in a dense \ac{TSDF}-based map while limiting the growth of the map, based on probabilistic measures;
\item completely CPU-based implementation to allow for use on lightweight robotic platforms lacking GPU hardware;
\item the addition of threading infrastructure and an approximate fast integrator to the open-source TSDF library Voxblox.
\end{itemize}

\section{Related Work}
\label{sec:related_work}

There has been extensive work on \ac{SLAM} over the last three decades, of which visual \ac{SLAM} forms a significant part \cite{cadena2016past}. Dense 3D mapping from image data has also received considerable research focus, and in recent years there has been a surge in the number of works in this field. A plethora of systems have been developed, for which we give a brief review of the most relevant.

Dense mapping systems have employed a number of different representations of the environment, a choice which determines many properties of the resulting system. Engel et. al. \cite{engel2014lsd} represent the world as a collection of well localized depth frames produced by multi-view stereo. Whelan et. al. \cite{whelan2015elasticfusion} represent the world as a collection of surfels in 3D space. Another approach is occupancy grids, which represent the world as a collection of occupancy probabilities stored over a voxel grid \cite{hornung2013octomap}, and have been applied successfully to robotic systems in the past \cite{burri2015real}. In this work we represent the observed world by maintaining a \ac{TSDF} over a discrete grid, an approach which has shown compelling results in recent years \cite{newcombe2011kinectfusion}. Such representations offer a systematic method for incremental fusion of noisy depth frames, provide high fidelity reconstructions, and make no assumptions about the structure of the environment. Recent works on sparse representations of \ac{TSDF}s have shown that these techniques can also be employed at larger scales \cite{niessner2013real}. Furthermore, while keyframe depth-maps and surfel-based representations produce very accurate reconstructions, these representations are challenging to use directly for other robotic tasks such as motion planning. The \ac{TSDF}-based volumetric representation is more conducive for such tasks, where free space and surface connectivity information are important \cite{wagner2014graph,lin2017autonomous}.


Most existing online reconstruction systems use frame-to-frame or frame-to-model alignment for tracking the sensor pose at each frame, before integrating depth data into the map \cite{newcombe2011kinectfusion,niessner2013real}. These kind of incremental systems inherently suffer from camera tracking drift, which can lead to dramatic distortions in the reconstructed environment \cite{niessner2013real}. There have been several works which aim to address the challenge of producing consistent dense maps, of which we review the most relevant.

Whelan et. al. \cite{whelan2012kintinuous} introduce a system which restricts the volume of active \ac{TSDF} reconstruction, converting data leaving this volume to a triangular mesh. To ensure consistency, the global mesh is deformed to reflect loop closures following a rigid-as-possible approach. ElasticFusion \cite{whelan2015elasticfusion} took a similar approach to maintaining consistency, achieved by bending a surfel-based map, which allowed for place revisiting. BundleFusion \cite{dai2016bundlefusion} enforced consistency by storing the history of integrated frames. Camera poses are globally optimized with each arriving frame and the global reconstruction updated, producing compelling globally consistent reconstructions. This method however faces scalability issues during on-line use.




Another approach to maintaining a consistent map is to represent the global map as a combination of submaps. The global map can then be computed as a function of these submaps only. This approach is not new, going back at least to the \emph{Atlas} framework \cite{bosse2003atlas} and DenseSLAM~\cite{nieto2006denseslam} where submapping was used to extend graph-based \ac{SLAM} to large scale environments. Several works have shown the efficacy of \ac{TSDF}-subvolume approaches for maintaining map consistency \cite{henry2013patch, kahler2016real, zhou2013elastic, wagner2014graph, fioraio2015large}. These works can be broadly separated into two categories; those that attempt to \emph{partition} space to minimize subvolume overlap, and those that do not partition space.

Partitioning the workspace means that the constructed map size grows linearly with the size of the observed environment, rather than time or trajectory length. Patch Volumes \cite{henry2013patch} divides the observed scene into ``map patches'', which are aligned with planar surfaces in the environment. In a similar approach K\"{a}hler et al. \cite{kahler2016real} attempt to minimize subvolume overlap. While space efficient, there are two disadvantages of pursuing an approach of subdivision. Firstly, when revisiting an existing subvolume, one must ensure the camera pose is consistent with that volume in order to integrate the information, which in practice is difficult. Secondly, both \cite{henry2013patch} and \cite{kahler2016real} require ray casting into several subvolumes simultaneously to perform camera tracking, significantly increasing the computational requirements of the system.

By contrast, several systems \cite{zhou2013elastic}, \cite{wagner2014graph}, \cite{fioraio2015large} make no attempt to partition space or to control subvolume overlap. The authors in \cite{zhou2013elastic} build reconstructions as a composition of mesh fragments and produce compelling results. Both systems however, are aimed at producing high quality reconstructions in post processing, leading to high computation times, making these approaches inappropriate for real-time robotics applications. The authors in \cite{wagner2014graph} and \cite{fioraio2015large} represent the observed world as a number of potentially overlapping \ac{TSDF} volumes and show this to be effective for maintaining map consistency. We extend this approach. The main drawback of the existing works in this direction is that the maps they produce grow linearly with the trajectory length \cite{wagner2014graph}, or with time \cite{fioraio2015large}. Our proposed approach differs from these systems in a number of technical aspects (see section \refsec{sec:overview}), however the main scientific novelty of our system is to address the scalability problem by introducing a systematic approach to limiting map growth.

\begin{figure*}[!ht]
    \centering
  \includegraphics[width=0.65\textwidth]{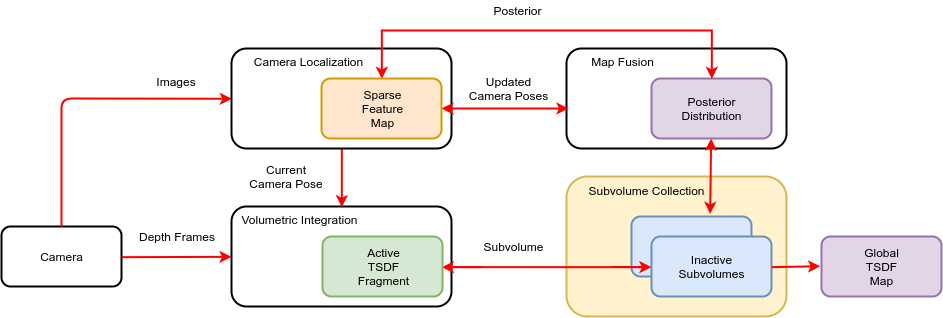}
    \caption{An overview of the proposed mapping system.}
    \label{fig:overview}
\end{figure*}

To summarize, the existing approaches which address the challenge of maintaining consistency in dense maps either: a) maintain a map in a form which requires further processing, for example point/surfel clouds or keyframe depth maps; b) make fusing new information into previously mapped locations difficult for example mesh-based representations; c) use offline or computationally expensive techniques for global optimization of the reconstruction; d) build maps which grow in size without bound, even in a bounded size environment. In constrast to existing work, the proposed system maintains the map in a form suitable for use in robotics, allows seemless place revisiting, and limits the growth of maps online.

\section{Problem Statement}
\label{sec:problem}

Given a sequence of images $\{I^i\}_{i=1}^M$ and associated depth maps $\{D^i\}_{i=1}^M$, we aim to build a dense 3D map of an observed scene. The camera coordinate frames $\{C^i\}_{i=1}^M$ are parameterized as rigid transformations with respect to a global reference frame $G$ as $\mat{T}_{GC^i} \in \text{SE}(3)$. Central to our approach is the use of a \ac{TSDF} for implicit surface representation, which we denote as the function $F: \mathbb{R}^3 \to (d, w, \vec{c})$ mapping 3D points to a tuple, consisting of $d$ the distance to the nearest observed surface, $w$ a weighting/confidence measure, and $\vec{c}$ the observed color \cite{newcombe2011kinectfusion}. As is common in recent reconstruction systems \cite{niessner2013real}, \cite{oleynikova2017voxblox}, we store this function as a collection of sparse samples over a discrete uniformly-spaced voxel grid. 

If the set of poses $\{\mat{T}_{GC^i}\}_{i=1}^M$ is well determined at the time of image capture, existing techniques \cite{newcombe2011kinectfusion}, \cite{niessner2013real}, \cite{oleynikova2017voxblox} can be used to construct a global \ac{TSDF} map. We however, focus on the case where camera frames are not well \emph{globally} localized at the time of capture, for example due to accumulated drift or delayed loop closure. 


\section{System Description}
\label{sec:overview}

The proposed system builds on the TSDF mapping techniques first presented in \cite{newcombe2011kinectfusion} and later expanded by a number of works \cite{oleynikova2017voxblox, whelan2012kintinuous, niessner2013real}. An overview of our system is shown in \reffig{fig:overview} and consists of modules for camera tracking/localization (\refsec{sec:sparse}), volumetric integration (\refsec{sec:reconstruction}), map fusion (\refsec{sec:maintenance}), and subvolume fusion (\refsec{sec:fusion}). We start with a description of our map respresentation as a subvolume collection.


\subsection{\ac{TSDF} Sub-volumes}
\label{sec:subvolumes}

When fusing observation data into a single map, inconsistencies can be caused by integrating data from a poorly localized sensor. Particularly in the case of dense maps, such errors tend to be difficult to recover from because correlations between observation and localization information are usually lost in order to ensure mapping remains efficient. We pursue a map structure which remains consistent \emph{by construction}. We represent the scene as a collection of locally consistent sub-volumes. However, we make no attempt to \emph{partition} the global space, allowing for a one-to-many relationship between the world and the map. This is advantageous for maintaining consistency as multiple (potentially conflicting) hypotheses about the structure of the environment are able to remain separate in the map until conflicts can be disambiguated, which can be arbitrarily delayed.

Each \ac{TSDF} subvolume represents a small section of the environment and the total map is represented by the collection of all such subvolumes, such that
\begin{equation}
\Pi = \{\pi^p\}_{p=1}^N 
\end{equation}
where $N$ is the number of currently constructed subvolumes. Each subvolume $\pi^p$ has a local coordinate system $M^p$ associated with it, parameterized by $\mat{T}_{GM^p} \in \text{SE}(3)$. The 3 operations possible on the subvolume collection are 1) addition of a new subvolume (\refsec{sec:reconstruction}) 2) modification of a subvolume's pose (\refsec{sec:maintenance}), or 3) destruction of a subvolume by fusion of its content with another subvolume (\refsec{sec:fusion}).





\subsection{Camera Localization}
\label{sec:sparse}

We utilize sparse, feature-based SLAM for providing the current sensor pose as well as updates to past poses. Typical \ac{TSDF} systems have tended to rely on depth image registration for camera tracking, however in the context of this work sparse, feature-based localization offers two advantages: 1) modern sparse systems run efficiently on a CPU allowing their application to lightweight robotic platforms (see \refsec{sec:results_platform}), and 2) a sparse feature map, in which correlations between localization information and the environment are maintained, allows for probability-based decision making (see \refsec{sec:maintenance}).

For this work we make use of ORB-SLAM2 \cite{mur2016orb} with the modifications proposed in \refsec{sec:maintenance}. In the vein of the de-facto standard in modern visual slam, ORB-SLAM2 maintains a graph comprised of a set of keyframes and landmarks, with keyframe poses $\mathcal{K}$ and landmark positions $\mathcal{L}$ respectively. Camera tracking and mapping then are formulated as non-linear least squares optimizations of feature re-projection errors on the image plane,
\begin{align}
\begin{split}
\{\mathcal{K}^*_{\mathcal{O}}, \mathcal{L}^*_{\mathcal{O}}\}
&=
\argmin_{\mathcal{K}_{\mathcal{O}}, \mathcal{L}_{\mathcal{O}}}
\sum_{k \in \mathbb{K}}
\sum_{l \in \mathbb{L}_k}
\rho
\left( \left\lVert
\mathbf{e}_{k,l} )
\right\rVert_{\boldsymbol{\Sigma}}^2 \right)
\\
\mathbf{e}_{k,l}
&=
\vec{l}^l_{(\cdot)} - \boldsymbol{\pi}_{(\cdot)}(\mat{T}_{K^kG}(\mat{L}^{l})),
\label{eq:dwo}
\end{split}
\end{align}
where the sets $\mathcal{L}_{\mathcal{O}} \subset \mathcal{L}$ and $\mathcal{K}_{\mathcal{O}} \subset \mathcal{K}$ are the optimized landmarks and keyframes respectively, and are subsets of the full map. The extent of these subsets determines the level of mapping fidelity. 
The set of keyframe indices is denoted $\mathbb{K}$ and the set of landmark indices observed by keyframe $k$ is denoted $\mathbb{L}_k$. The cost function $\rho$ is the Huber robust cost and $\boldsymbol{\Sigma}$ is the covariance matrix associated with the keypoint. The position of the $l^\text{th}$ landmark is $\mat{L}^l \in \mathbb{R}^3$ and $\mat{T}_{K^kG} \in SE(3)$ parameterizes the pose of the $k^\text{th}$ keyframe. ORB-SLAM2 allows for both monocular and stereo (depth) feature points $\vec{l}^l_{(\cdot)}$, such that $\vec{l}^l_m \in \mathbb{R}^2$ and $\vec{l}^l_s \in \mathbb{R}^3$, and $\boldsymbol{\pi}_{(\cdot)}$ is either the stereo or monocular projection function. In this work we limit ourselves to the scenario where \emph{some} 3D observations are available to constrain the scale of the estimated map.

\subsection{Volumetric Integration}
\label{sec:reconstruction}

Incoming image data is added to the collection by integrating arriving frames $I^i$ and $D^i$ into the currently active subvolume $\pi_A \in \Pi$. We first transform the estimated sensor pose, supplied by camera tracking, into the active subvolume frame $M_A$
\begin{equation}
\mat{T}_{M^{A}C^i}
=
(\mat{T}_{GM^{A}})^{-1} \oplus \mat{T}_{GC^i},
\end{equation}
where $\oplus$ denotes composition on SE(3). For integration of depth-frame data we use the open source \ac{TSDF} toolbox, \textit{voxblox}, introduced in \cite{oleynikova2017voxblox} and extend it to provide interfaces for submapping.

Periodically, a new subvolume is created, marked as active and the last active subvolume is transferred to the collection, growing the map. Subvolumes are rigidly attached to the first keyframe after their creation such that $M^p \equiv C^k$ where $k$ is the image index of the $k\text{th}$ keyframe. We assume that data \emph{within} a subvolume is consistent, and therefore our approach is to generate new subvolumes early and often, relying on the map fusion module to ameliorate the scalability cost of doing so. We generate a new active subvolume after either: the number of keyframes contributing to a subvolume reaches a maximum value, or there is significant change in the sparse map which is likely to cause a jump in the camera tracking.

The original voxblox system employed a computationally demanding frame integration method which was not suitable for real-time dense mapping on computationally-limited platforms. Also, unlike most TSDF libraries, voxblox operates on generic pointclouds rather than depth images. It also explicitly maps all free-space. This means that many of the strategies other libraries use for increasing runtime performance cannot be applied (chiefly image-space preprocessing and voxel projection). To overcome these issues we implement a new multi-threaded ``fast'' integration approach\footnote{\url{https://github.com/ethz-asl/voxblox}}.

The fast integrator operates by terminating the ray casting early in two cases where the structure of an area is likely to have already been captured by other rays. The first case governs the maximum density of the starting points. Each point is first inserted into a ``starting location'' set that has twice the resolution of the standard voxel map. If a point already occupies this starting location the point is discarded. The second termination condition is to check if a ray is updating voxels that have already been updated by other rays in the same frame. This check is done by inserting the point into an ``observed voxel'' set. If a ray attempts to update two voxels in a row that have already been updated, it is deemed to be adding minimal information and is terminated. As the rays draw together as they are near the camera location this acts to terminate many rays shortly after the surface while still ensuring the vast majority of freespace voxels are allocated and at least partially updated. The speed of this approach depends on the speed of insertions into the sets. They are implemented as fixed-size, one-element-per-bucket, hash-sets, where collisions are resolved by discarding the old value. This allows synchronous lock-free insertions and reads using atomic compare-and-swap instructions. These sets are cleared at the end of each frame integration.

The time to integrate a frame depends on the structure of the input pointcloud, thus some frames may be significantly more expensive then others. To ensure real-time performance is always maintained, a maximum integration time was implemented that will terminate integration early if a time budget is exceeded.


\subsection{Map Fusion}
\label{sec:maintenance}

In this section we propose an approach for maintaining consistency of the map while also limiting its growth. Following a loop closure, camera localization provides an updated set of keyframe poses resulting from global optimization of the sparse map \refeq{eq:dwo}. The representation of the environment within each subvolume, as relative to its base frame, means that the collection is corrected simply by updating subvolume coordinate frames $\{M^p\}_{p=1}^N$ with their updated poses.

The system described thus far will produce a collection of subvolumes which increases in size linearly with trajectory length, meaning that during operation, even in bounded size environments, the map grows without bound. This limits the practical applicability of the proposed system. However, when revisiting a place, the collection is likely to contain multiple, potentially redundant, views of the same area. Our approach to limiting map growth is to identify these redundant views and fuse them, where doing so has a high probability of producing consistent results. We utilize the information contained in the sparse feature map to govern this process.

To identify subvolumes which contain redundant views, we propose to use a modified version of the \emph{covisibility graph}~\cite{mei2010closing}. During construction of the map we associate each keyframe $K^i$ with the subvolume to which it contributed. We then build a weighted graph $G = (V, E)$ where the vertex set $V$ represents the subvolumes $\{\pi^p\}_{p=1}^N$, and the edge set $E$, with associated weights $W$, represent landmark covisibility information. Formally, vertex $i$ and $j$ have an edge of weight 
\begin{equation}
  W_{ij} = \vert \mathbb{L}_i \cap \mathbb{L}_j \vert,
\end{equation}
where $\mathbb{L}_i$ and $\mathbb{L}_j$ are the sets of landmark indices observed by keyframes contributing to subvolumes $\pi_i$ and $\pi_j$, and here $\vert \cdot \vert$ indicates the cardinality of the set. Subvolume pairs connected with an edge of high weight in this graph are likely to have viewed the same area, and are therefore passed as candidate pairs to the next stage.

At this stage we have a list of subvolume pairs which contain views of similar sections of the environment. However, the estimated poses of these subvolumes may contain significant localization error. Naive fusing of such pairs is likely to produce inconsistent results, and we would therefore rather maintain these views separately (as they currently exist) in the collection. Thus, before fusing subvolume pairs we determine a measure of the accuracy of their relative localization. Formally, given a candidate pair $(\pi_i, \pi_j)$, with base-frames $M_i$ and $M_j$, and associated with keyframes $K_i$ and $K_j$, we define $q(i,j)$ the quality measure of their relative localization as
\begin{equation}
  q(i,j) = 1 / \Vert \mat{\Sigma}_{i \vert j} \Vert,
\end{equation}
where $\Sigma_{i \vert j}$ is the covariance matrix associated with the conditional distribution $P(K_i \vert K_j)$. For this work the norm $\Vert \cdot \Vert$ is the 2-norm, which is proportional to the volume of an ellipsoid of constant probability density defined by $\Sigma_{i \vert j}$. Subvolumes pairs meeting a minimum value of $q$ are fused together. In the remainder of this section we discuss how $\mat{\Sigma}_{i \vert j}$ is determined.

\subsubsection{Extraction of the Subvolume Conditional Covariance}

Given an initial guess of the optimization parameters $\theta = (\mathcal{K}, \mathcal{X})$, solving bundle adjustment proceeds by iteratively linearizing the graph and solving the linear system,
\begin{equation}
  \mat{H} \theta = \mat{b},
  \label{eq:ba_linear}
\end{equation}
where the typically sparse matrix $\mat{H}$ and vector $\mat{b}$ are determined from the measurements, their Jacobians, covariances, and linearization point (see \cite{kummerle2011g} for a detailed exposition). At convergence we are left with an approximation of the information matrix
\begin{equation}
  \mat{\mathcal{I}} \equiv \mat{H}^*
\end{equation}
where $\mat{H}^*$ results from linearizing the graph at the convergence point $\mathbf{\theta}^*$. Inverting the (typically large) information matrix $\mathcal{I}$ to get covariance matrix $\mat{\Sigma}$ associated with the posterior distribution is computationally expensive and does not scale to practical problems. However, determining $\mat{\Sigma}_{i \vert j}$ for each subvolume-pair requires very few blocks of the full matrix $\mat{\Sigma}$; two diagonal blocks and a single off-diagonal block. We aim to perform the minimum computation required to extract these blocks, and proceed as follows.

We first marginalize out the landmarks from the graph using the so-called \emph{Schur Compliment trick} to form the reduced camera matrix, here denoted $\mat{\mathcal{I}}_{\mathcal{P}}$. This reduction can be performed very efficiently for problems of the size typically encountered in \ac{SLAM} (see \cite{agarwal2010bundle} full exposition on this method). From here we utilize the dynamic programming approach based on Cholesky decomposition of the information matrix introduced in \cite{kaess2009covariance}. This algorithm, calculates a  requested covariance matrix element $\sigma_{i,j}$, as a function of a collection of other covariance matrix entries which follow a sparsity pattern which is a subset of the non-zero entries of the Cholesky factor $\mat{R}$.


We propose a modification to the original algorithm which increases its speed for calculating a small number of covariance matrix elements. The speed of the original algorithm depends predominantly on the number of intermediate entries required during computation of the desired element $\sigma_{i,j}$. The number of these elements needed is a product of the ordering of optimization variables prior to Cholesky decomposition. Using an ordering aimed at reducing overall fill-in, such as \ac{AMD} \cite{davis2008user}, can lead to widely varying and potentially large computation times when extracting a very small set of elements. We propose then to re-order the matrix $\mathcal{I}_{p}$ prior to Cholesky decomposition for \emph{each} keyframe candidate pair for which we need covariance entries. By reordering with each pair, we are free to force the blocks associated with $\Sigma_{i \vert j}$ to appear last in the Cholesky factor. This leads to a vast reduction in the number of intermediate elements computed, and a corresponding increase in speed. To achieve this reodring we use a constrained version of \ac{AMD} \cite{davis2008user}.

We pay for this ordering method in two ways. Firstly, constraining the position of some blocks of $\mathcal{I}$ is likely to increase the overall fill-in of the Cholesky factor. Secondly, we perform a decomposition of $\mathcal{I}_{p}$ for \emph{each} subvolume-pair candidate, instead of just once for all candidates, as is done in the original algorithm \cite{kaess2009covariance}. In our testing, despite these drawbacks, there was a substantial benefit in taking the proposed approach.



\subsection{Subvolume Fusion}
\label{sec:fusion}

We frequently are required to fuse subvolumes, both at the request of the map maintainance module, and for visualizing the current state of the subvolume collection. Given two subvolumes to be fused $\pi_i$ and $\pi_j$ with a reference frames $M_i$ and $M_j$ the transformation between them and global map is then found 
\begin{equation}
\mat{T}_{M^iM^j} = (\mat{T}_{GM^i})^{-1} \oplus \mat{T}_{GM^j}.
\end{equation}
We then transform the voxels in $\pi_j$, allocate new voxels in $\pi_i$ where needed, and integrate their contributions using trilinear interpolation. For visualization of the whole manifold we repeat this process multiple times, fusing all subvolumes into $\pi_1$.

\section{Results}
\label{sec:results}

The results presented in this section aim to validate our main contribution; a system for maintaining consistency while limiting map growth. We therefore evaluate the performance of our system in terms of the accuracy of reconstructed surfaces, and the size of the produced maps on several datasets. Finally, we validate our claim of applicability to lightweight robotic platforms by demonstrating the system in use on an \ac{MAV}.

\subsection{Reconstruction Quality}
\label{sec:results_reconstruction}

\begin{figure}[t] 
    \centering
    \begin{subfigure}[b]{0.40\columnwidth}
    \includegraphics[trim={0cm 0cm 0cm 0cm},clip,width=\columnwidth]{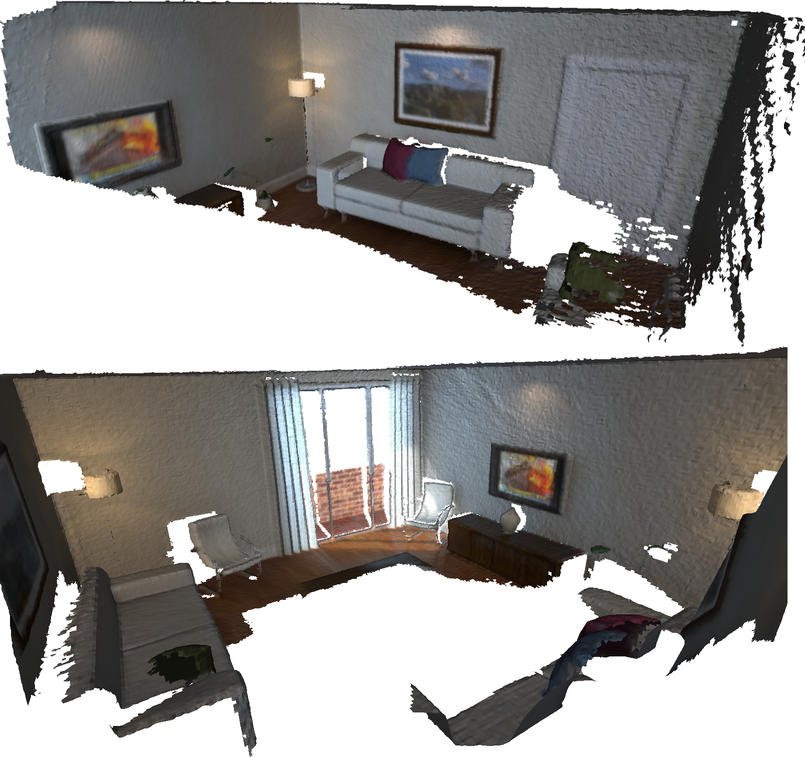}
        \caption{}
        \label{fig:kt_mesh}
    \end{subfigure}
    \begin{subfigure}[b]{0.40\columnwidth}
    \includegraphics[trim={0cm 0cm 0cm 0cm},clip,width=\columnwidth]{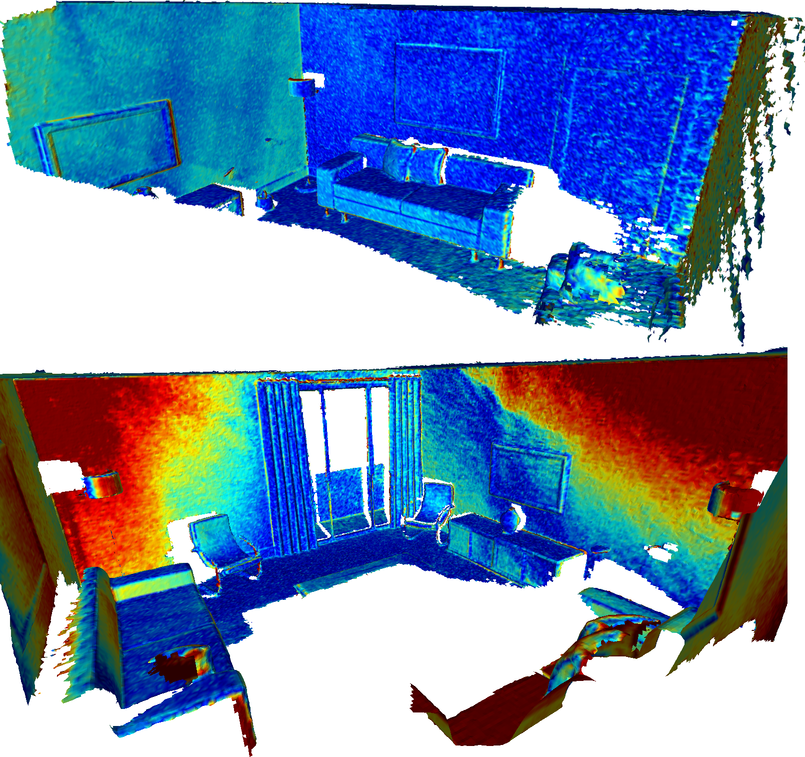}
        \caption{}
        \label{fig:kt_error}
    \end{subfigure}
    \begin{subfigure}[b]{0.15\columnwidth}
    \includegraphics[trim={-0.25cm -1.0cm 0cm 0cm},clip,width=\columnwidth]{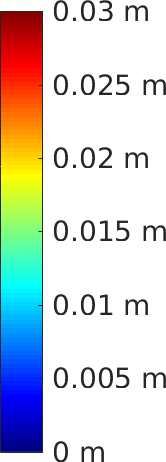}
    \end{subfigure}
    \caption{Reconstructions from 2 \emph{living\_room} trajectories of the ICL-NUIM dataset \cite{handa:etal:ICRA2014} with (\subref{fig:kt_mesh}) observed color and (\subref{fig:kt_error}) colored by error. The median error across all 4 trajectories is 0.015\,m for a voxel size of 0.02\,m.}
    \label{fig:icl_nuim_rgbd}
\end{figure}

We evaluate the surface reconstruction accuracy of our system using the publicly available ICL-NUIM dataset \cite{handa:etal:ICRA2014}. This dataset provides RGB and depth images captured by a camera moving through a synthetic room. In addition, ground truth geometry is provided for evaluation of reconstructed surfaces. We reconstruct the scene using the proposed system as well the baseline system ElasticFusion \cite{whelan2015elasticfusion}, a top performing reconstruction system which makes heavy use of GPU-based computation. Reconstructions are performed on a desktop PC with an Intel Core i7-7820X CPU at 3.60\,GHz, and a Nvidia Titan Xp GPU (donated by the Nvidia Corporation), which is used only by ElasticFusion. \reffig{fig:icl_nuim_rgbd} shows reconstructed meshes produced by the proposed system for the kt0 and kt1 \emph{living\_room} sequences.

We align the surface reconstructions with the ground truth geometry and calculate the \ac{RMSE} between mesh vertices (or surfel centers) and the closest point on the ground-truth surface. In addition, we evaluate the surface reconstruction produced by voxblox~\cite{oleynikova2017voxblox} employing our fast integration method and supplied with ground truth camera poses. The performance of this combination is an indication of the minimum achievable error using the frame integration method proposed in \refsec{sec:reconstruction}. Both the proposed method and voxblox use a voxel size of 0.02\,m. \reftab{tab:icl_nuim_rgbd} shows the results of these evaluations.

Our system achieves a median \ac{RMSE} error across the four sequences of 0.015\,m. Given our voxel size of 0.02\,m, an error less than this amount indicates the system is performing well. Furthermore, our system performs only slightly worse than the same integration system utilizing ground truth poses, which scores 0.013\,m. The proposed method scores slightly worse than ElasticFusion on sequences \textit{kt0}, \textit{kt1}, and \textit{kt2}, which scores a median error of 0.001\,m. Given our design choices, which enable use on systems lacking a GPU this performance is to be expected. As we will show in \refsec{sec:results_platform} one advantage of our system is that it can be used on lightweight robotic platforms. Sequence \textit{kt3} is the only sequence containing a loop closure. On this sequence the proposed system outperforms ElasticFusion. While one cannot judge a systems performance on a single dataset, this is an indication that our mechanism for maintaining map consistency is performing as expected.

\begin{table}[t]
\centering
\ra{1.3}
\begin{tabular}{@{}rrrrr|r@{}}\toprule
\phantom{abc} & \multicolumn{5}{c}{RMSE (m)} \\
\cmidrule{2-6}
System & kt0 & kt1 & kt2 & kt3 & median \\
\midrule
ElasticFusion & 0.006 & 0.009 & 0.010 & 0.048 & 0.001 \\
Voxblox (GT Poses) & 0.010 & 0.017 & 0.014 & 0.011 & 0.013 \\
Ours & 0.011 & 0.024 & 0.016 & 0.013 & 0.015 \\
\bottomrule
\end{tabular}
\caption{Comparison of surface reconstruction \ac{RMSE} for the proposed system, ElasticFusion \cite{whelan2015elasticfusion} and voxblox \cite{oleynikova2017voxblox} supplied with ground truth poses evaluated of the ICL-NUIM dataset \cite{handa:etal:ICRA2014}. The results indicate that our CPU-based system achieves surface reconstruction performance comparable to a state-of-the-art approach utilizing a GPU.}
\label{tab:icl_nuim_rgbd}
\end{table}

\subsection{Map Maintenance}
\label{sec:results_maintainance}

We evaluate the performance of the map-maintenance module for limiting map-growth, and analyze the effects of subvolume fusion on reconstruction accuracy. To obtain input data we use CARLA \cite{Dosovitskiy17}, an open-source driving simulator. The simulator produces photo realistic scenes of a car driving in a synthetic city, providing sensor output with realistic (non-ideal) camera effects, (noiseless) depth-maps, and ground-truth camera poses.

We evaluate our system using data from drives through two synthetic cities which include significant exploration, wide-baseline loop closures, and place revisiting. Our system is evaluated in two configurations: subvolume fusion turned ON and turned OFF. For comparison we evaluate a naive approach that uses voxblox for reconstruction and ORB-SLAM2 for camera tracking, but has no method to correct the dense map following loop closure. Lastly, we reconstruct the scene using ground truth camera poses and voxblox which represents a measurement of the minimum achievable error using our fast frame integration method and voxel size (0.5\,m). In all configurations tracking and frame integration occurs at 10\,Hz.

Carla does not provide ground-truth geometry for their maps, so we reconstruct our own ``ground-truth'' geometry with which we evaluate surface reconstructions. We generate a fine-grained reconstruction with voxblox using ground-truth poses, in a non real-time configuration which uses 0.25\,m voxels and the more costly frame integration method described in \cite{oleynikova2017voxblox}. This is an approximate measure of ground-truth structure, but for the remainder of this section we will assume errors are predominantly produced by the tested reconstruction systems.

\begin{figure}[t]
    \centering
    \begin{subfigure}[b]{0.20\textwidth}
    \includegraphics[trim={0cm 0cm 0cm 0cm},clip,width=\columnwidth]{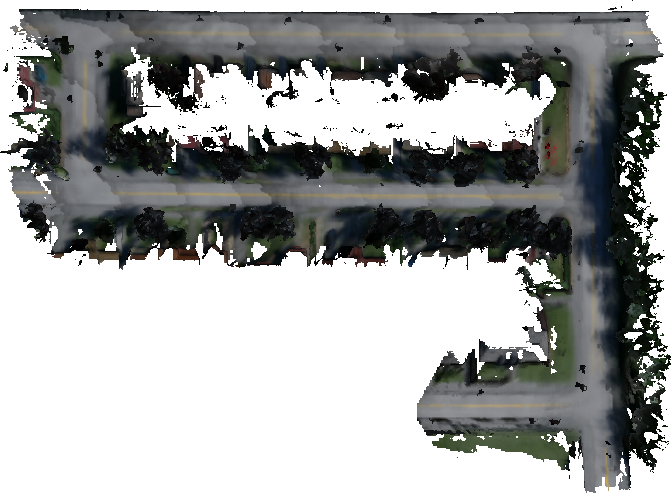}
        \caption{}
        \label{fig:l0_bird}
    \end{subfigure}
    \begin{subfigure}[b]{0.25\textwidth}
    \includegraphics[trim={-2.5cm 0cm 0cm 0cm},clip,width=\columnwidth]{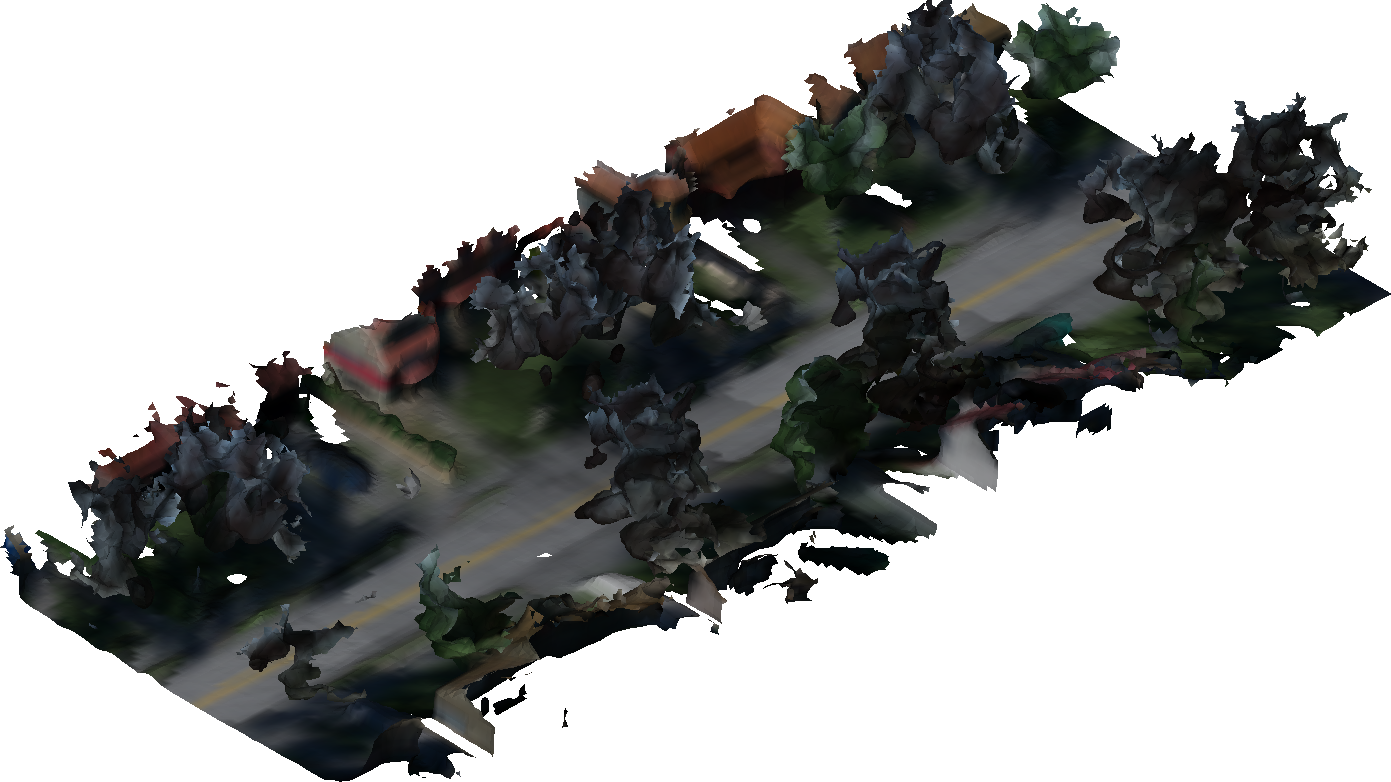}
        \caption{}
        \label{fig:l0_close}
    \end{subfigure}
    \begin{subfigure}[b]{0.20\textwidth}
    \includegraphics[trim={0cm 0cm 0cm 0cm},clip,width=\columnwidth]{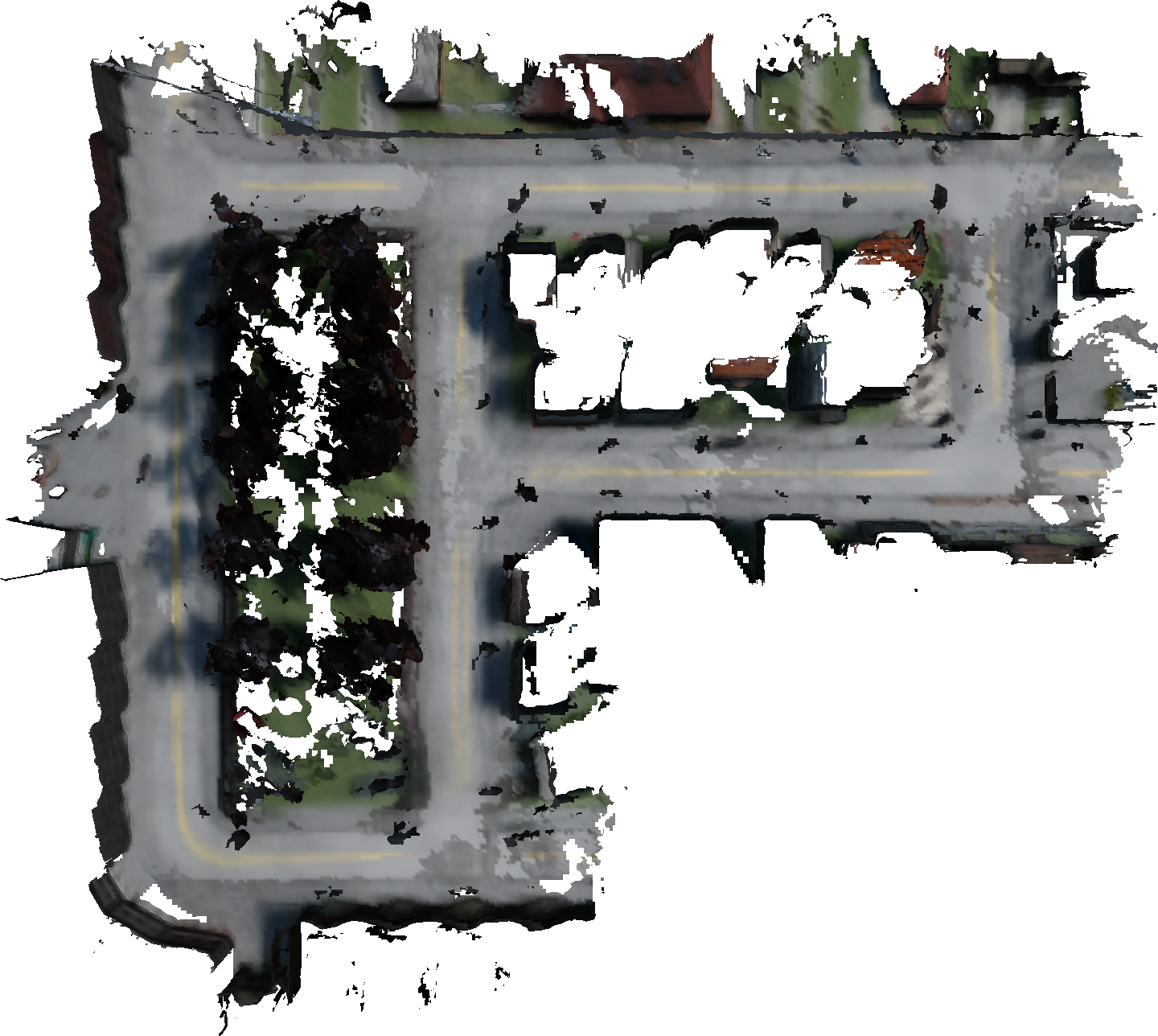}
        \caption{}
        \label{fig:l1_bird}
    \end{subfigure}
    \begin{subfigure}[b]{0.25\textwidth}
    \includegraphics[trim={0cm 0cm 0cm 0cm},clip,width=\columnwidth]{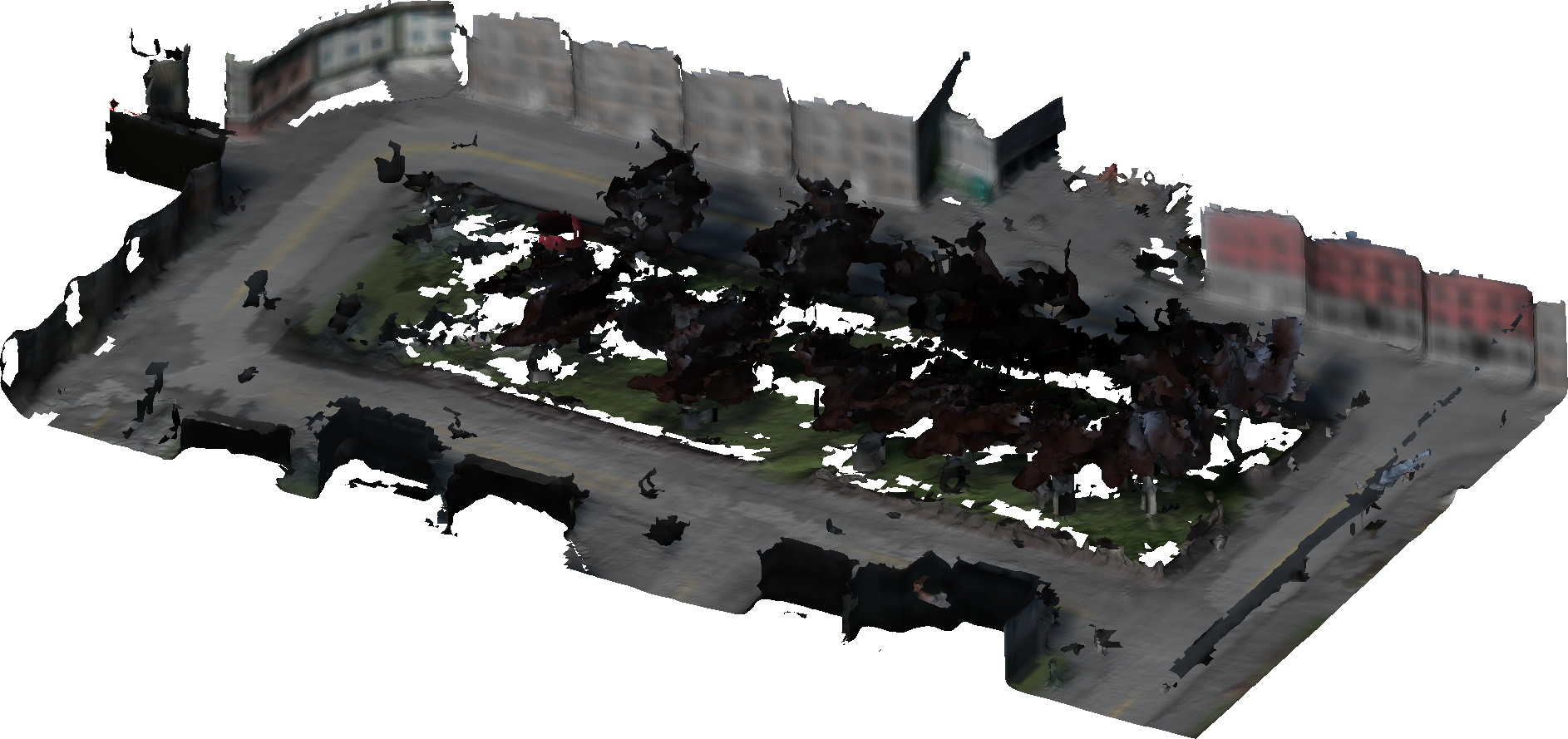}
        \caption{}
        \label{fig:l1_close}
    \end{subfigure}
    \caption{Reconstructions produced by our system with data generated by a car driving through 2 synthetic cities \cite{Dosovitskiy17}. Sub-figures (\subref{fig:l0_bird}) and (\subref{fig:l1_bird}) show bird's-eye views of the reconstructed areas, and (\subref{fig:l0_close}) and (\subref{fig:l1_close}) are closes ups of areas of these maps showing a tree-lined street and a square with house fronts visible respectively.}
    \label{fig:carla}
\end{figure}

Reconstructions from our system for two drives through different cities are shown in \reffig{fig:carla}. Visible in the reconstructions is the structure of the synthetic cities, showing streets, houses, and trees. For quantitative evaluation we compute RMSE reconstruction error (computed as in \refsec{sec:results_reconstruction}) and final map size, in terms of allocated blocks. The results of this evaluation for each system configuration are shown in \reftab{tab:carla}.

\begin{figure}[t]
    \centering
    \begin{subfigure}[b]{0.20\textwidth}
    \includegraphics[trim={0cm 0cm 0cm 0cm},clip,width=\columnwidth]{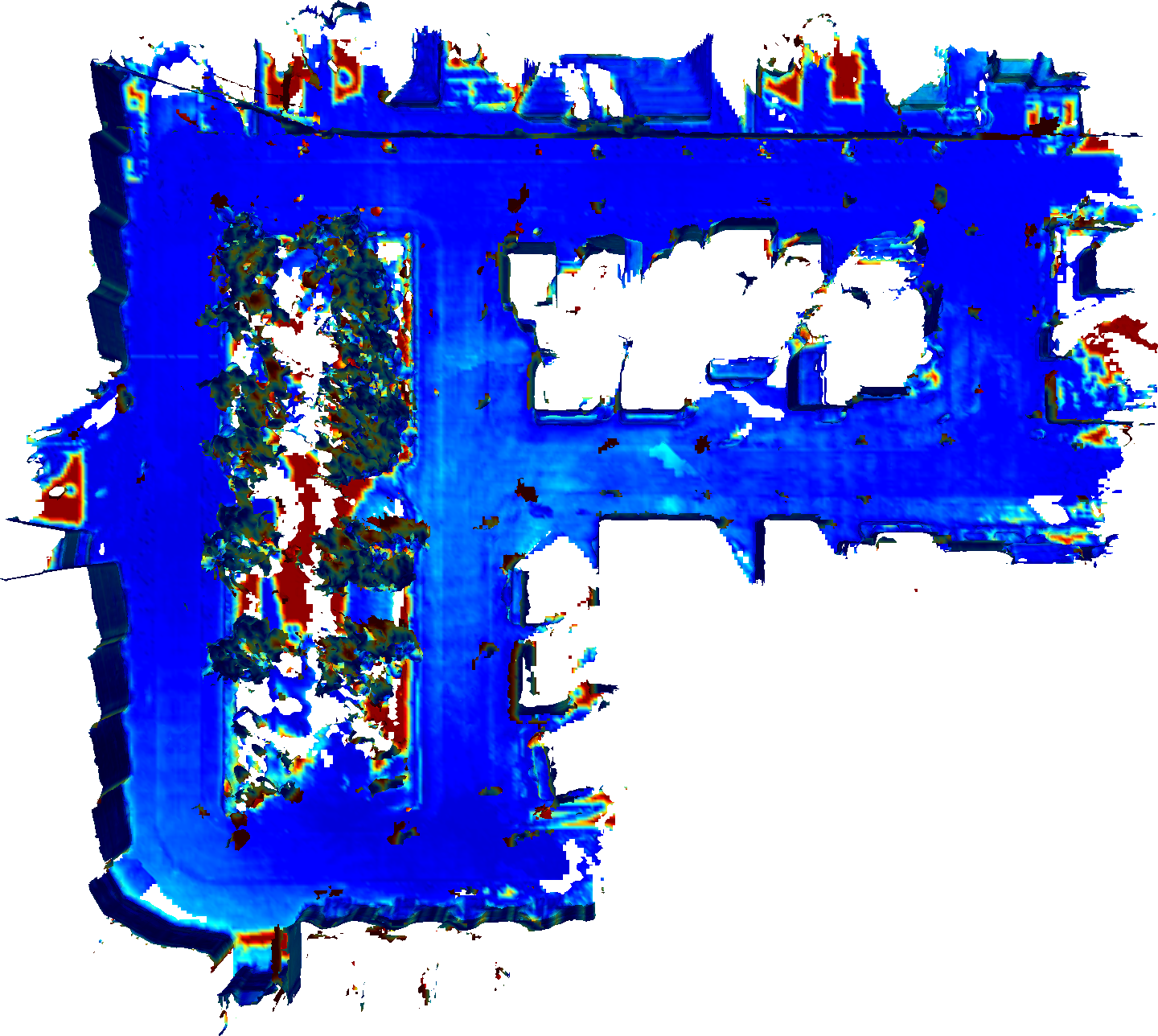}
        \caption{}
        \label{fig:l1_error_ours}
    \end{subfigure}
    \begin{subfigure}[b]{0.20\textwidth}
    \includegraphics[trim={0cm 0cm 0cm 0cm},clip,width=\columnwidth]{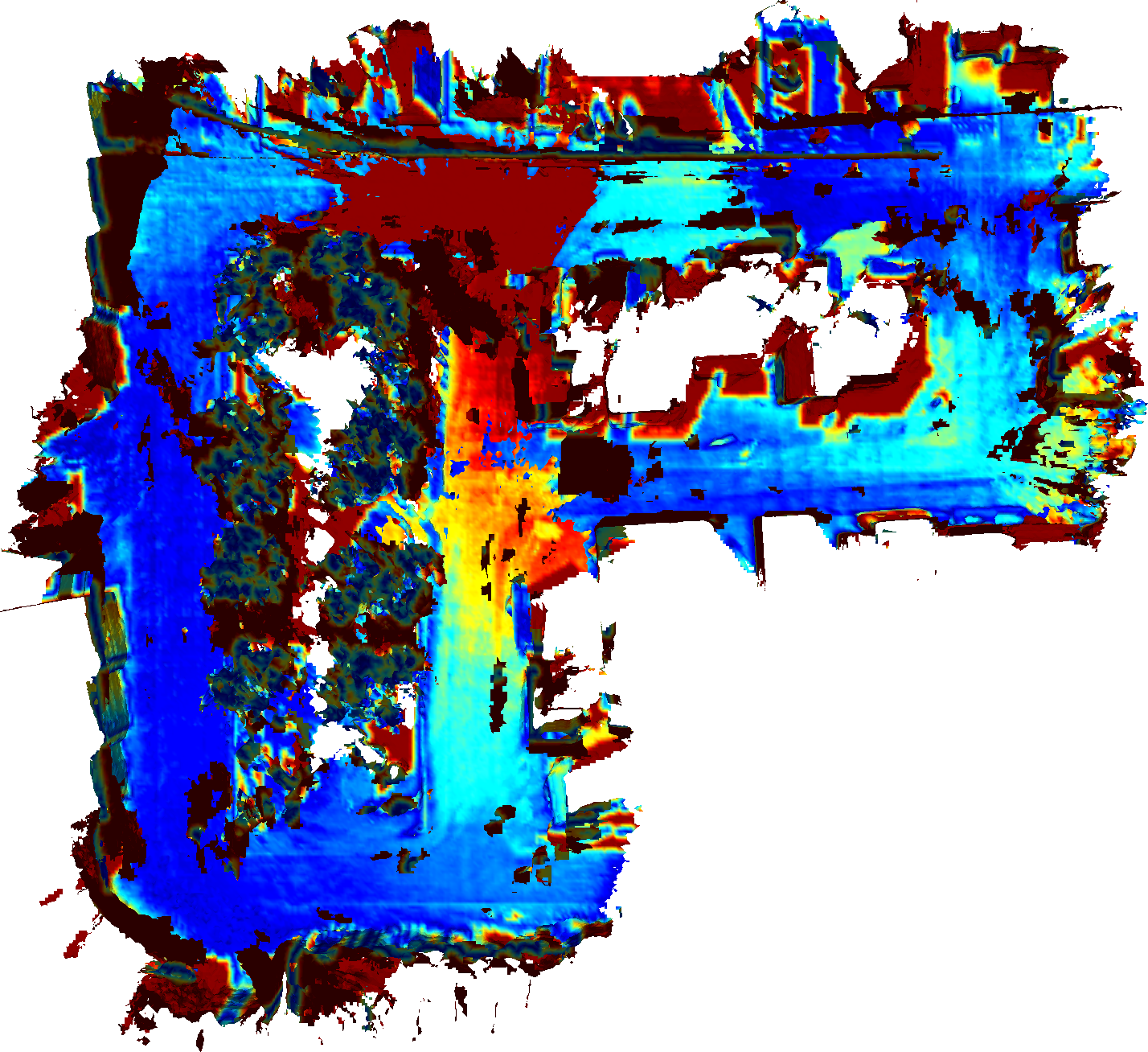}
        \caption{}
        \label{fig:l1_error_orb}
    \end{subfigure}
    \begin{subfigure}[b]{0.06\textwidth}
    \includegraphics[trim={-0.25cm -1.0cm 0cm 0cm},clip,width=\columnwidth]{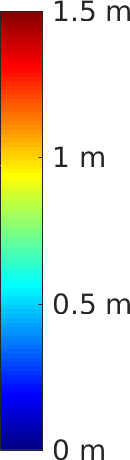}
    \end{subfigure}
    \caption{Reconstructions of an urban environment produced by the proposed system (\subref{fig:l1_error_ours}) and a naive approach not explicitly addressing map consistency (\subref{fig:l1_error_orb}). The reconstructions are colored by surface error. The area of high error in the center of reconstruction (\subref{fig:l1_error_orb}) is the site of several wide-baseline loop closures, indicating that error accumulates in these areas, if consistency is not explicitly maintained.}
    \label{fig:carla_error}
\end{figure}

For both trajectories, the results show that the lowest error is achieved by the system with access to ground truth poses as expected. This error score of median 0.60\,m is approximately the dimension of one voxel, which is reasonable given the approximations described in \refsec{sec:reconstruction}. The system using voxblox with ORB-SLAM2 for camera tracking yields a median score of 2.09\,m, 348\% of the error score of the system using ground-truth poses. \reffig{fig:l1_error_orb} shows the reconstructed mesh colored by error. The areas of high error in the center of the map are at the location of two wide-baseline loop-closures. It is expected that in these areas, where drift has accumulated maximally, error is the highest. The median reconstruction errors of our system in the two tested configurations (0.68\, and 0.72\,m) approach the error of the system with access to ground-truth poses, with increases in the error of 13\% and 20\% respectively, indicating a significant improvement over the naive approach.

The subvolume fusion system is effective in reducing the map size to 55\% in \textit{l0} and 68\% on \textit{l1} with respect to the system with no method of map fusion. The amount of compression achievable is heavily dependent on the path taken, as frequent revisiting will lead to areas of significant redundancy and produce good candidates for fusion. Similarly, the threshold in terms of the quality measure $q$ at which fusion occurs will determine the number of subvolumes fused, and the incurred cost in terms of reconstruction error. There is a tradeoff to be made here, earlier fusion resulting from a lower $q$ threshold will result in smaller maps, however will likely incur higher costs in terms of reconstruction accuracy, as submaps are fused before being optimally localized.

\begin{table}[t]
\centering
\ra{1.0}
\begin{tabular}{@{}rrr|rrr@{}}\toprule
\phantom{abc} & \multicolumn{3}{c}{RMSE (m)} & \multicolumn{2}{c}{Size (blocks)} \\
\cmidrule(lr){2-4} \cmidrule(lr){5-6} 
System & l0 & l1 & med. & l0 & l1 \\
\midrule
Voxblox (GT Poses)       & 0.52 & 0.70 & 0.60 & 6315 & 4561 \\
Voxblox (ORB-SLAM Poses) & 2.12 & 2.06 & 2.09 & 7205 & 5240 \\
Ours (subvolume fusion OFF)  & 0.59 & 0.77 & 0.68 & 28908 & 17856 \\
Ours (subvolume fusion ON)   & 0.66 & 0.77 & 0.72 & 15873 & 12220 \\
\bottomrule
\end{tabular}
\caption{Comparison of surface reconstruction \ac{RMSE} and final map size, in number of allocated voxel blocks, for the proposed system in two configurations (subvolume fusion ON and OFF), and voxblox \cite{oleynikova2017voxblox} supplied with either ground-truth or ORB-SLAM2 \cite{mur2016orb} estimated poses. Data is collected during two drives (\textit{l0} and \textit{l1}) through two synthetic cities using CARLA \cite{Dosovitskiy17}.}
\label{tab:carla}
\end{table}

\subsection{Platform Results}
\label{sec:results_platform}

To demonstrate the applicability of the proposed framework to a lightweight robotic platform we reconstruct an industrial environment using a lightweight \ac{MAV}. The hexacopter is based on a DJI F550 frame equipped with a \ac{VI}-sensor~\cite{nikolic2014synchronized} and an Intel RealSense D415 Depth Camera. A px4 autopilot performs low-level attitude stabilization, while high-level computation, including the proposed approach, was run entirely on-board and in real-time on an Intel NUC Core i7-7567U. For our purposes, the \ac{VI}-sensor provided tightly synchronized stereo images from a pair of global shutter cameras, and is used for camera tracking at 10\,Hz. The D415 provided colored pointclouds which are integrated into the map at the same rate. Calibration between the tracked camera and depth camera frames is performed offline using the \textit{Kalibr} toolbox\footnote{\url{https://github.com/ethz-asl/kalibr}}.

The modifications made to the voxblox integrator described in \refsec{sec:reconstruction} reduce the average time required to integrate depth data in this environment from 93ms to 25ms per frame, allowing the dense mapping to run in real-time while still providing sufficient CPU time for the other components of the proposed pipeline, as well as other software components related to control of the \ac{MAV}.

Figure~\ref{fig:cover} shows reconstructions from two flights, \emph{f0} and \emph{f1}, as well as the mesh produced during from \emph{f0} colored by subvolume membership and a view from the \ac{VI} sensor during \emph{f1}. During flight \emph{f0} the \ac{MAV} performed a 251\,s trajectory which included two wide-baseline loop-closures and subsequent activation of the bundle adjustment and map fusion pipeline. In total 18 subvolumes were fused in flight \emph{f0}, from a total of 34 created. The view from the onboard \ac{VI} sensor shows pipe structures which are visible in the reconstruction \reffig{fig:cover_pipes}.

\section{Conclusion}
\label{sec:label}


In this paper we proposed a novel system for producing consistent dense 3D maps, which includes a systematic approach for limiting map growth, and runs in real time on a CPU. Our system is based on maintaining a collection of overlapping \ac{TSDF} subvolumes which together constitute the global map. To limit map growth we propose a method for fusing redundant, well-localized subvolumes. Our evaluations on public, simulated and self-collected datasets, show that the proposed system is able to correct for significant inconsistencies that stem from imperfect camera localization. In our evaluation of the system on the ICL-NUIM benchmark, our system achieves a median \ac{RMSE} of 0.015\,m, which approaches the error score of a state-of-the-art reconstruction system requiring a high-power GPU. In a simulated urban environment we show the efficacy of our system for maintaining map consistency versus a naive approach, and that our subvolume fusion system is effective in significantly reducing the size of the resulting map. Finally, we deploy the proposed system to a \ac{MAV} and generate a consistent reconstruction over two flights through an industrial area, demonstrating the system's applicability to lightweight robotic platforms. The ability to produce consistent and scalable dense maps, represents a step towards allowing robotic platforms to maintain detailed 3D maps in large scale environments.


\bibliographystyle{ieeetr}
\bibliography{bibliography}

\end{document}